\documentclass[a4paper,fleqn]{cas-dc}
\usepackage[authoryear,longnamesfirst]{natbib}
\usepackage{subfigure}
\begin{document}

\shorttitle{Maximize the Foot Clearance for a Hopping Robotic Leg Considering Motor Saturation} 
\title[mode = title]{Maximize the Foot Clearance for a Hopping Robotic Leg Considering Motor Saturation}
  
\shortauthors{Juntong Su et al.}  

\author[1]{Juntong Su}[style=chinese]
  
\author[1]{Bingchen Jin}[style=chinese]

\author[1]{Shusheng Ye}[style=chinese]

\author[3]{Lecheng Ruan}[style=chinese]

\author[1,2]{Caiming Sun}[style=chinese]

\author[1,2]{Ning Ding}[style=chinese]

\author[4]{Yili Fu}[style=chinese]
  
\author[1,2]{Jianwen Luo}[style=chinese]
\cormark[1]

\address[1]{Shenzhen Institute of Artificial Intelligence and Robotics for Society (AIRS), Shenzhen 518172, China.}
\address[2]{Institute of Robotics and Intelligent Manufacturing (IRIM), The Chinese University of Hong Kong (CUHK), Shenzhen 518172, China.}
\address[3]{Beijing Institute of General Artificial Intelligence (BIGAI), Beijing, 100000, China}
\address[4]{State Key Laboratory of Robotics and System, Harbin Institute of Technology, Harbin, 150000, China.}

\tnotetext[1]{This work was supported in part by National Natural Science Foundation of China under Grant 51905251.}

\cortext[cor1]{Corresponding author: Jianwen Luo. Email: jamesluo@cuhk.edu.cn}


\maketitle
 
 \begin{abstract}[S U M M A R Y]
  A hopping leg, no matter in legged animals or humans, usually behaves like a spring during the periodic hopping. Hopping like a spring is efficient and without the requirement of complicated control algorithms. Position and force control are two main methods to realize such a spring-like behaviour. The position control usually consumes the torque resources to ensure the position accuracy and compensate the tracking errors. In comparison, the force control strategy is able to maintain a high elasticity. Currently, the position and force control both leads to the discount of motor saturation ratio as well as the bandwidth of the control system, and thus attenuates the performance of the actuator. To augment the performance, this letter proposes a motor saturation strategy based on the force control to maximize the output torque of the actuator and realize the continuous hopping motion with natural dynamics. The proposed strategy is able to maximize the saturation ratio of motor and thus maximize the foot clearance of the single leg. The dynamics of the two-mass model is utilized to increase the force bandwidth and the performance of the actuator. A single leg with two degrees of freedom is designed as the experiment platform. The actuator consists of a powerful electric motor, a harmonic gear and encoder. The effectiveness of this method is verified through simulations and experiments using a robotic leg actuated by powerful high reduction ratio actuators.
 \end{abstract}
\section{Introduction}	
Legged locomotion is one of the most vibrant areas. Its versatile motion ability enables many behaviours similar to humans and animals. Although a lot of knowledge has been obtained about the bio-inspired legged locomotion such as hopping \cite{6522811, 8202172, 9028203}, walking \cite{donghyun2020, luo3dwalking, luo_oussama}, running \cite{6094504, 7139828, 7527657}, etc., competence in these gaits is yet comparable to that of the biological counterparts. One of the reasons is that these actuators do not deliver sufficient performance and corresponding control strategy to improve these behaviours. To achieve a high-performance actuator, several  electromagnetic (EM) actuators have been proposed to this end. EM actuators mainly fall in three categories: direct-drive actuator (or proprioceptive force control actuator \cite{7827048}, quasi-direct-drive \cite{8202172, 6880316}), series elastic actuator (SEA) \cite{803231, 9250572, 8016601} and high-ratio geared motor with torque sensor \cite{7758092}. 
\begin{figure}[t]
\centering
\setlength{\abovecaptionskip}{0.1 cm}
\setlength{\belowcaptionskip}{3 cm}
\includegraphics[width = 3.3 in]{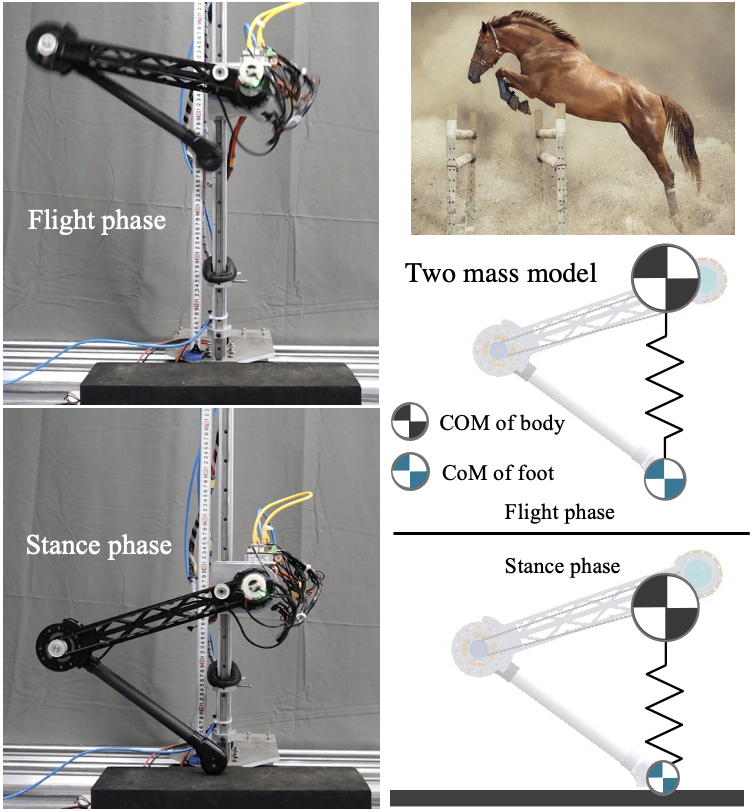}
\caption{Maximize the foot height of a single hopping robotic leg. In nature, legged animals fully retract legs to maximize the feet clearance as to hop over a high obstacle as shown in the right top subfigure.}
\label{fig:concept}
\end{figure}
Due to the low gear ratio of the direct-drive actuator, the friction is ignorable and the control algorithm can be simplified. However, the disadvantage is that the direct-drive actuator has lower power density (power per unit mass) as well as torque density (torque per unit mass) and thus is not able to provide enough load capacity.

The Ground Reaction Forces (GRFs) that is equivalent to the 2.6 to 3 times of the body weight are needed during the high speed locomotion \cite{walter2007ground}. Therefore, the high torque density is desirable for the dynamic robotic systems \cite{6386252}. To leverage the torque density, harmonic gears have been widely used in various legged robots due to the compact size, high reduction ratio and low weight of the harmonic gears. The power density and torque density of the actuator equipped with harmonic reducer are able to be leveraged significantly compared with the direct-drive and quasi-direct-drive. There are several paradigms of well-known legged robots that rely on harmonic gears to increase the output torque of the motor, including the humanoid robot HRP-2P \cite{hirukawa2005human}, Asimo \cite{1041641}, and ANYmal \cite{7758092}. These robots demonstrated impressive locomotion ability, which proves that the harmonic gear is effective in improving the load capacity of the actuator. It is noteworthy that 
SEA improves the backdrivability with elastic transmission  \cite{525827} \cite{4783213} \cite{jianwen2018}. The SEA uses a mechanical component to provide the elasticity and to facilitate active compliance control. It can achieve impressive force control as well as impact tolerance, however, the force control performance of the SEA is considered to have limitations due to its elasticity, and thus not comparable to the rigid actuators in terms of bandwidth \cite{803231, 7579567}.

This study proposes the concept of motor saturation ratio and maximizes foot clearance of single leg in hopping experiments. To this end, the dynamics and kinematic of the single leg based on two-mass model are analyzed in Section \uppercase\expandafter{\romannumeral2}. Then, both the control methods and simulation verification are elaborated, and the relationship between desired torque and motor saturation in force control is introduced in Section \uppercase\expandafter{\romannumeral3}. In \uppercase\expandafter{\romannumeral4}, experimental verification is conducted to demonstrate that the proposed strategy realizes hopping with natural dynamics while maximizing the foot clearance.

The contributions in this letter lie in the following twofold:

1) A motor saturation strategy based on two-mass model is proposed to maximize the foot clearance of a hopping leg.

2) A high-power actuator consisting of EM and harmonic gear is adopted in the hopping leg platform for experiment validation of motor saturation strategy.

The rest of this letter is organized as follows. Related work is reviewed in Section II. Two-mass model for a hopping leg is analyzed in Section III. Control method considering motor saturation and simulation results are presented in Section IV. Section V demonstrates the experiment results. This line of research is concluded in Section VI. 
\section{Related Work}
Bezier curve has been extensively adopted for the trajectory planning of the joint position in legged locomotion \cite{8260563, luo3dwalking}. However it is not appropriate for hopping since it is not obtained by coupled dynamics. To make the control law more consistent with the hopping characteristics, the time response of dynamics based on the two-mass hopping leg model is proposed \cite{9028203}. However, when using two-mass model, part of the actuator's capacity is used to ensure the position accuracy and compensate for the inertia errors, especially during the start moment. When position control is implemented in the hopping task, the saturation ratio (defined in section \uppercase\expandafter{\romannumeral3}) is not able to be fully increased. Force control is an effective method to achieve the active impedance. Virtual model control (VMC) was adopted on the quadruped robot HyQ such that HyQ's leg dynamics is sharped equivalent to a virtual spring from the hip joint to the foot. This achieves the active compliance of the robot leg \cite{6696541}. Impedance control is also adopted on another single leg actuated by hydraulic \cite{BA2020103704}. These studies are based on hydraulic actuation which is powerful but energy-inefficient though. However, two aspects of the impedance control based on single mass model still needs to be optimized. First, the impact force at the landing point is still large, especially for the actuator equipped with harmonic reducer. A strategy based on two-mass model for the optimal landing is proposed, which is able to minimize the impact \cite{9028203}. Second, the strategy used above is not capable of fully taking advantage of the performance of the actuator due to the negligence of motor saturation. As the frequency of hopping increases, the motor controlled by these strategies becomes unstable and dangerous.

In the traditional algorithm, the spring stiffness of the two-mass model generally remains constant, so as to obtain the hopping locomotion with a fixed law \cite{9028203}. Although the constant stiffness makes the control system simpler, it also limits the force bandwidth of the control system. Force control with motor saturation can maximize the spring stiffness of the two-mass model, so that the actuator can output the maximum force in the stance phase to achieve the maximum foot clearance. At the same time, this approach is able to mitigate the spring stiffness at the landing moment to protect the motor from damage.

In our study, to achieve the hopping task with foot clearance maximized on an robot leg, a force control based on two-mass model has been proposed. The overall strategy in this letter aims at maximizing the foot clearance for a single leg, as illustrated in Fig. \ref{fig:concept}. The controller is designed to decouple the coupled dynamics of the leg in the task space, and thus to control the hopping task (force control) and the swing motion (position control) separately. In the force control, the boundary values of torque based on the motor saturation model, are introduced to increase the saturation ratio of motor and force bandwidth. The desired torque of the knee joint is maximized in the stance phase to realize the maximum value of foot clearance. Due to the inherent characteristics of the two-mass model, the single leg can rapidly recover to a stable posture and achieve continuous hopping.
\section{Two-mass Model for a Hopping Leg}
\begin{figure}[t]
\centering
\includegraphics[width  = 3.2 in]{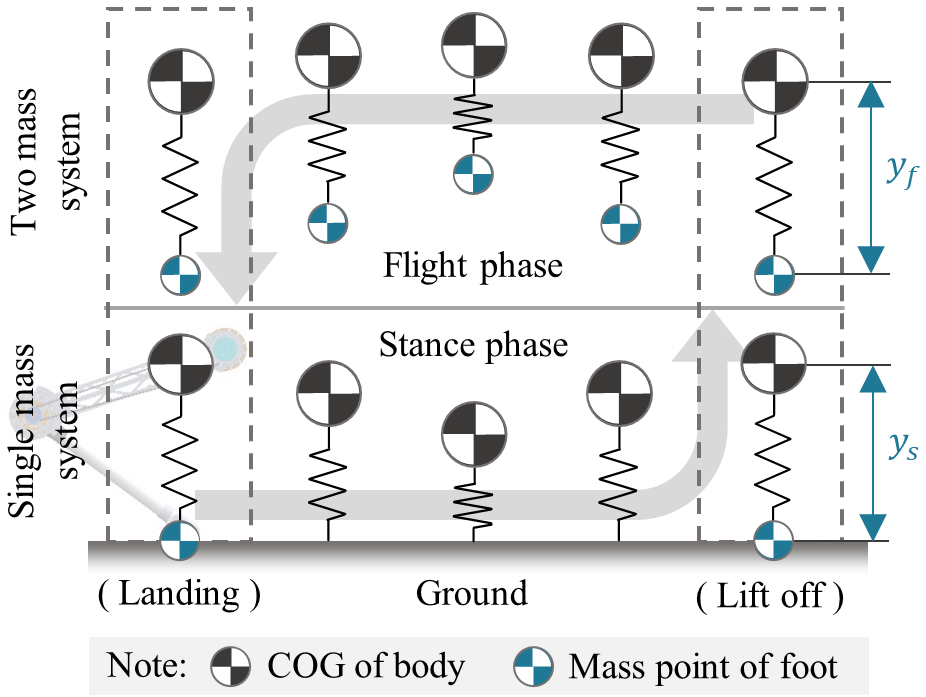}
\caption{The complete hopping cycle of the two-mass model.}
\label{fig:twomass_1}
\end{figure}

To solve for the locomotion of the single leg, the detailed dynamics and kinematic of the two-mass model is examined in this section. As shown in Fig. \ref{fig:twomass_1}, the two-mass model consists of two masses which are the body and the foot that is connected via a spring. The two-mass model can explain the large impact of the landing as the foot is considered a mass point, which is more suitable for single leg than single mass model for landing moment.

\subsection{Dynamics Analysis of the Two-mass Model}
The dynamics of the two-mass model in stance phase can be be simplified to single mass model as follows:
\begin{equation}
	m\ddot y_s \left( t \right) = mg-k_s\left(y_s\left(t\right)-y_{s\cdot neu} \right),
	\label{eq:dyn_stance}
\end{equation}
where $y_s (t)$ is the time response for the distance from the COG of body to the foot end in stance phase and $k_s$ is the stiffness of the spring. $y_{s\cdot neu}$ indicates the neutral position of the spring system. $m$ is the mass of the body and $g$ is the gravity.

In flight phase, the dynamics of the two-mass model is switch to:
\begin{equation}
\begin{cases}
	m\ddot y_m \left( t \right) = mg-k_{f \cdot m} y_{f\cdot m}\left(t\right) \\
	m_e\ddot y_e \left( t \right) = m_e g-k_{f \cdot e} y_{f\cdot e}\left(t\right)
	\label{eq:dyn_flight}
\end{cases},
\end{equation}
where $\ddot y_m \left(t\right)$ and $\ddot y_e \left(t\right)$ are the absolute acceleration equation respectively for the COG of the body and the mass point of foot. $m_e$ is the mass of the foot. $k_{f\cdot m}$ and $k_{f\cdot e}$ are the relative stiffness of the two-mass model that are defined as $k_{f\cdot m}=k_s \frac{m+m_e}{m_e}$ and $k_{f\cdot e}=k_s \frac{m+m_e}{m}$. $y_{f\cdot m}\left(t\right)$ and $y_{f\cdot e}\left(t\right)$ are the time response for the distance from the COG of body and the mass point of foot to the neutral position of the two-mass model in flight phase.

The switching conditions are as follows:
\begin{itemize}
    \item Lift: $y_s \left(t_{lo}\right)=\frac{m_e g}{k_s}+y_{s\cdot neu}$, where $t_{lo}$ is the time for lift off. 
    \item Landing: $\dot y_e \left(t_{ld}\right) = 0$ and $y_{f\cdot e}\left(t_{ld}\right)=\frac{m_e g}{k_{f\cdot e}}$, where $t_{ld}$ is the time for landing.
\end{itemize}

\begin{figure}[h]
\centering
\includegraphics[width = 3.2 in]{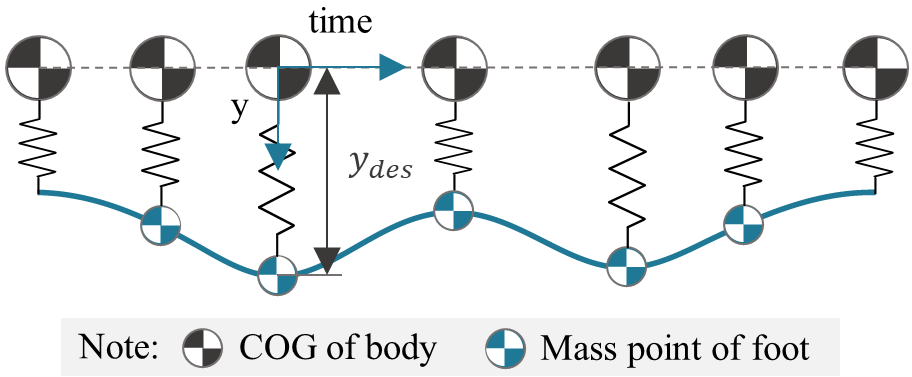}
\caption{Example of the time response of the hopping dynamics.}
\label{fig:twomass_2}
\end{figure}

\vspace{-3mm}
\subsection{Kinematics Analysis of Two-mass Model}
The trajectories of the body are calculated based on the dynamic model in equation (\ref{eq:dyn_stance}) and (\ref{eq:dyn_flight}). The time response of the stance dynamics is given as:
\begin{equation}
	y_s\left(t\right)=\left(C_{amp}+\frac{mg}{k_s}\right)  \cos{\left(\sqrt{\frac{k_s}{m}}t+\pi\right)}+y_{s\cdot neu},
	\label{eq:kin_stance}
\end{equation}
where $C_{amp}$ is the amplitude of the two-mass model in the stance phase.

The time response of the flight dynamics is given as:
\begin{equation}
\begin{split}
	&y_f\left(t\right)=\\
	&\sqrt{\frac{m\dot y_s ^2 \left(t_{lo}\right)m_e ^3}{k_s \left(m+m_e\right)^3}+\frac{y_s ^2 \left(t_{lo}\right)m_e ^2}{\left(m+m_e\right)^2}}\cos{\left(\sqrt{\frac{k_s\left(m+m_e\right)}{m\cdot m_e}}t\right)}+\\
	&\sqrt{\frac{m\dot y_s ^2 \left(t_{lo}\right)m ^3}{k_s \left(m+m_e\right)^3}+\frac{y_s ^2 \left(t_{lo}\right)m ^2}{\left(m+m_e\right)^2}}\cos{\left(\sqrt{\frac{k_s\left(m+m_e\right)}{m\cdot m_e}}t\right)}+\\
	&y_{s\cdot neu}.
	\label{eq:kin_flight}
\end{split}
\end{equation}

The effective time range of the above equation is $t_{f\cdot s}<t<t_{f\cdot e}$.
\begin{equation}
\begin{split}
	t_{f\cdot s}&=-\sqrt{\frac{m \cdot m_e}{k_s \left(m+m_e\right)}}\\
	&\arccos{\left(\frac{y_s \left(t_{lo}\right)m_e}{m+m_e}\sqrt{\frac{m \cdot m_e ^3 \dot y_s ^2\left(t_{lo}\right)}{k_s \left(m+m_e\right)^3}+\frac{m_e ^2 y_s ^2 \left(t_{lo}\right)}{\left(m+m_e\right)^2}}\right)},\\
	t_{f\cdot e}&=t_{f\cdot s}+2\pi \sqrt{\frac{m\cdot m_e}{k_s\left(m+m_e\right)}}.
	\label{eq:kin_tf}
\end{split}
\end{equation}

The switching time can be obtained as follows:
\begin{equation}
    \begin{cases}
	t_{lo}=&\sqrt{\frac{m}{k_s}}\left(\arccos{\frac{m_e g}{k_s\left(C_{amp}+\frac{mg}{k_s}\right)}}-\pi\right) \\
	t_{ld}=&\sqrt{\frac{m}{k_s}}\left(\arccos{\frac{m_e g}{k_s\left(C_{amp}+\frac{mg}{k_s}\right)}}-\pi\right)+ \\
	&2\pi \sqrt{\frac{m\cdot m_e}{k_s \left(m+m_e\right)}}
	\end{cases},
	\label{eq:kin_tlold}
\end{equation}

The switching conditions are $y_s\left(t_{lo}\right)$ and $\dot y_s\left(t_{lo}\right)$. The period of continuous hopping is obtained:
\begin{equation}
	T=2 t_{lo}+2\pi\sqrt{\frac{m\cdot m_e}{k_s \left(m+m_e\right)}}.
	\label{eq:kin_tcycle}
\end{equation}

In order to compensate for friction and inertia errors, the position compensation coefficient is introduced:
\begin{equation}
\begin{split}
	&C_{com}\left(t\right)= \\
	&\begin{cases}
	1, & 0 \leq t < \frac{T}{2}\\
	0.5 C_{max}\cos{\left(\frac{4\pi}{T}t\right)} - 0.5 C_{max}+1, & \frac{T}{2} \leq t \leq T\\
	\end{cases},
	\label{eq:kin_ccom}
\end{split}
\end{equation}
where $C_{max}$ is the maximum of the position compensation coefficient.

The time response of the dynamic for the single-legged robot in a single cycle is deduced as follows:
\begin{equation}
\begin{split}
	&y_{des}\left(t\right)= \\
	&\begin{cases}
	C_{com}\left(t\right)y_s\left(t\right), & 0 \leq t < t_{lo}\; and \;t>\left(T-t_{lo}\right)\\
	C_{com}\left(t\right)y_f\left(t\right), & t_{lo} \leq t \leq \left(T-t_{lo}\right)\\
	\end{cases}.
	\label{eq:kin_ydes}
\end{split}
\end{equation}

Example of the time response of the dynamic for the two-mass model is shown in Fig. \ref{fig:twomass_2}.
\begin{figure}[t]
\centering
\setlength{\abovecaptionskip}{0.1 cm}
\setlength{\belowcaptionskip}{3cm}
\includegraphics[width = 3 in]{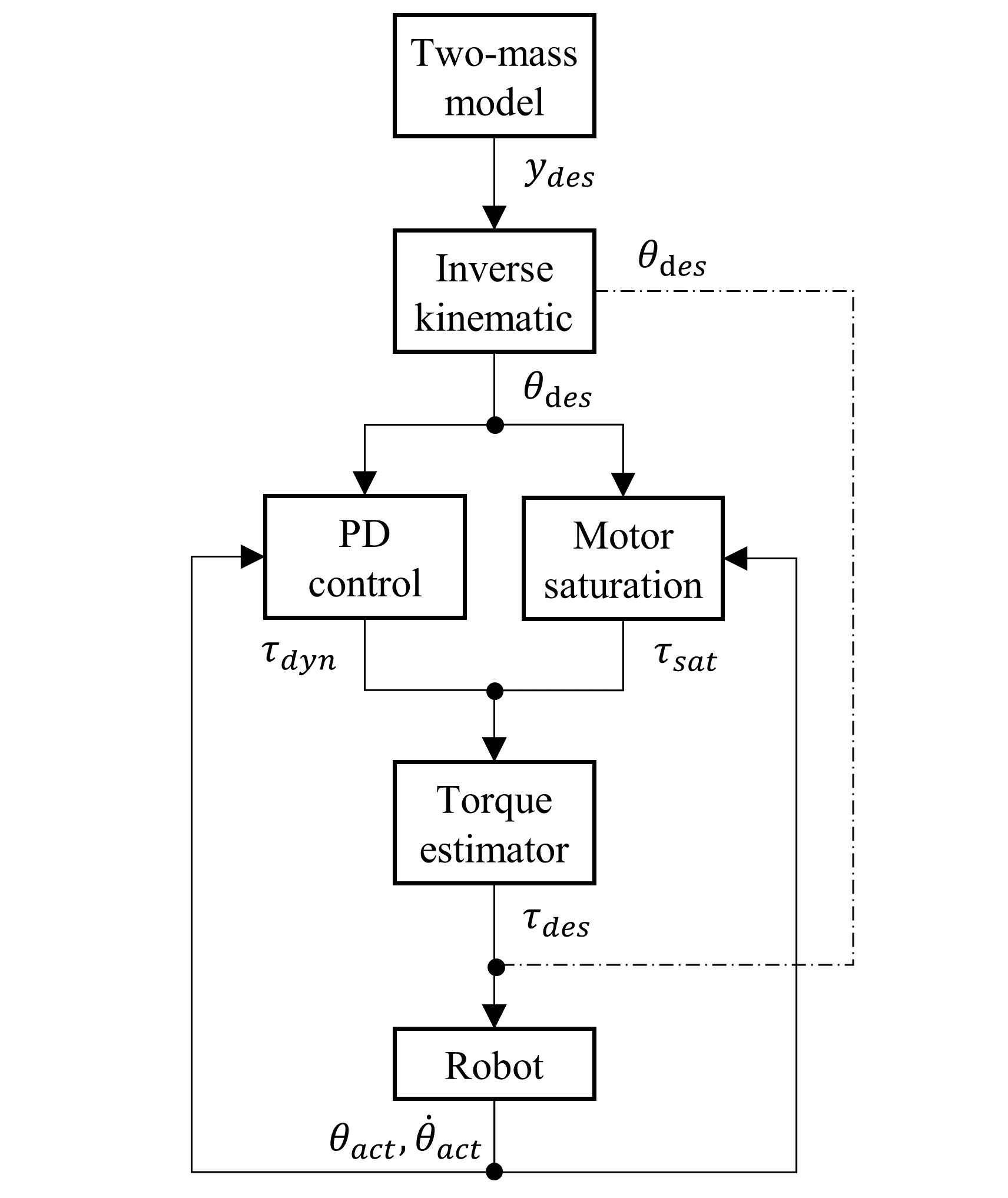}
\caption{Control diagram of force and position control (chain line).}
\label{fig:diagram}
\end{figure}

\section{Controller Design Considering Motor Saturation}
This section describes the control model of the single-legged robot in the hopping experiment. The control diagram of position control is shown in Fig. \ref{fig:diagram}.

\subsection{Force Control}
PD control is often used in force control. By combining the equation (\ref{eq:con_angdes}) and PD control, the desired torque of the actuators is obtained:
\begin{equation}
	\tau_{dyn}=-k_p\left(\theta_{act}-\theta_{des}\right)-k_d\dot\theta_{act},
	\label{eq:con_tdyn}
\end{equation}
where $\tau_{dyn}$ is the desired torque dynamics. $k_p$ is the proportional gain and $k_d$ is the derivative gain. $\theta_{act}$ indicates the actual angle of the actuators.

In order to better understand the effects of force and velocity saturation, we introduce a simple motor saturation model:
\begin{equation}
\begin{split}
	&\left|\tau_{sat\cdot motor}\right| \leq \begin{cases}
	\tau_{max}\left(1-\frac{\left|\dot \theta_{motor}\right|}{\omega_{max}}\right), & \left|\dot \theta_{motor}\right| \leq \omega_{max}\\
	0, & \left|\dot \theta_{motor}\right| > \omega_{max}\\
	\end{cases},
	\label{eq:con_tsatmotor}
\end{split}
\end{equation}
where $\tau_{max}$ and $\omega_{max}$ are the maximum torque and maximum angular velocity respectively for the motor. $\tau_{sat\cdot motor}$ is the saturation torque of motor. $\dot\theta_{motor}$ is the actual angular velocity of the motor.

The equation above introduces a linear back-EMF model where:
\begin{equation}
	\tau_{emf}=\frac{\tau_{max}}{\omega_{max}}\dot \theta_{act}.
	\label{eq:con_temf}
\end{equation}

The back-EMF can be thought of as an equivalent damping because it relates a loss in motor torque due to motor angular velocity.

When the motor is not pegged at saturation conditions, it still follows the back-EMF model \cite{803231}. To optimize hopping performance, we customized the saturation model of the motor.
\begin{equation}
\begin{split}
	&\tau_{sat} = \begin{cases}
	R\left(\tau_{max}\left(1-\frac{\left|R\dot \theta_{act}\right|}{\omega_{max}}\right)\right), & \left|R\dot \theta_{act}\right| \leq \omega_{max}\\
	0, & \left|R\dot \theta_{act}\right| > \omega_{max}\\
	\end{cases},
	\label{eq:con_tsat}
\end{split}
\end{equation}
where $\tau_{sat}$ is the output torque of actuator at the motor saturation point. $\dot\theta_{act}$ is the actual angular velocity of the actuator. $R$ is the reduction ratio of the joint.

The desired torque of the joints can be obtained as follows:
\begin{equation}
\begin{split}
	&\tau_{des} = \begin{cases}
	\tau_{dyn}, & \left|\tau_{dyn}\right| \leq \left|\tau_{sat}\right|\\
	sgn\left(\tau_{dyn}\right)\tau_{sat}, & \left|\tau_{dyn}\right| > \left|\tau_{sat}\right|\\
	\end{cases}.
	\label{eq:con_tdes}
\end{split}
\end{equation}

\begin{figure}[t]
\centering
\setlength{\abovecaptionskip}{0.2cm}
\setlength{\belowcaptionskip}{3cm}
\includegraphics[width = 2.8 in]{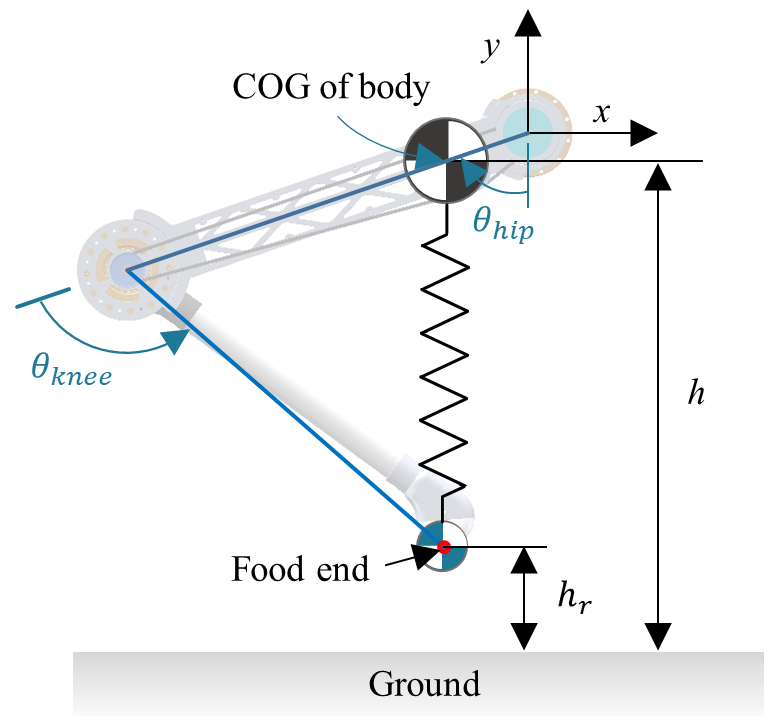}
\caption{Coordination of two-mass model with the single hopping leg.}
\label{fig:schematic}
\end{figure}

The foot clearance $h_r$, as shown in Fig. \ref{fig:schematic}, is an significant observed quantity to measure the performance of single leg. According to the conservation of energy, energy conversion equation of the single leg can be derived \cite{hutter2013starleth}:
\begin{equation}
\begin{aligned}
    	& mg\left(h_{r\cdot max}-h_{r\cdot init}\right)=\int ^{\theta _{lo}}_{\theta_{init}} \tau _{act} \left(\theta _{act}\right)\text d \theta _{act},
\end{aligned}
	\label{eq:con_energy}
\end{equation}
where $h_{r\cdot init}$ and $h_{r\cdot max}$ are the initial height and maximum height of the foot end. $\theta_{init}$ and $\theta_{lo}$ are the actual angle of the actuator to start and to lift off. $\tau _{act} \left(\theta _{act}\right)$ is the change law between $\tau _{act}$ and $\theta _{act}$.

Through the equation above, we can obtain the approach to increase the $h_{r\cdot max}$:
\begin{itemize}
    \item Increase the $\theta_{lo}$. In the actual model, $\theta_{lo}$ is related to many factors, such as $h_{r\cdot init}$, mechanical size, mass-distribution, impact, disturbance, etc. To simplify the experiment, we set the same $h_{r\cdot init}$ to trade off the $\theta_{lo}$.
    \item Increase the $\tau_{act}$.  It is an effective approach that is researched in our study.
\end{itemize}

To maximize the desired torque $\tau_{des}$, the proportional gain $k_p$ must be maximized ($k_p\geq 5400$).

\begin{figure}[tbp]	
\centering
\centering
\setlength{\abovecaptionskip}{0.0cm}
\subfigtopskip=0pt
\subfigbottomskip=6pt
\subfigcapskip=0pt
\vspace{-10pt}
\subfigure[]{
	\centering{\includegraphics[width = 3.0 in]{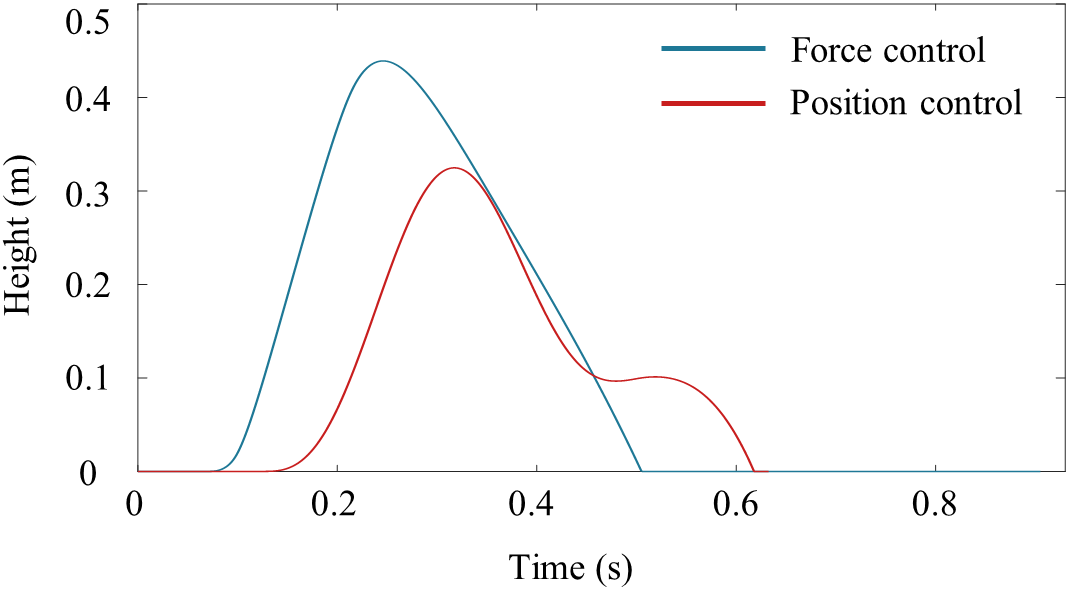}} 	
}
\subfigure[]{
	\centering{\includegraphics[width = 3.0 in]{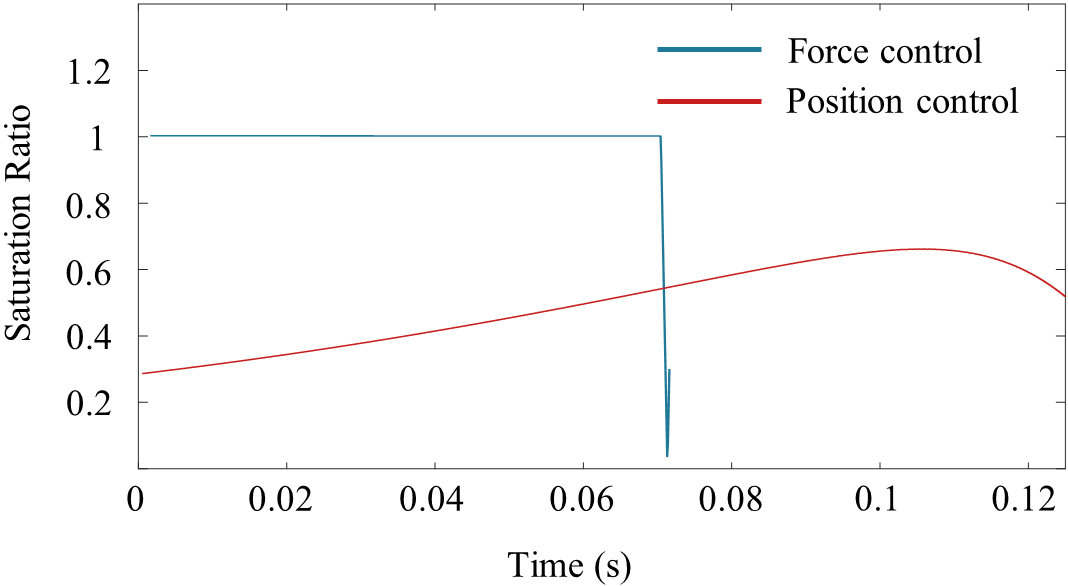}}
}
\subfigure[]{
\centering{\includegraphics[width  = 3.0 in]{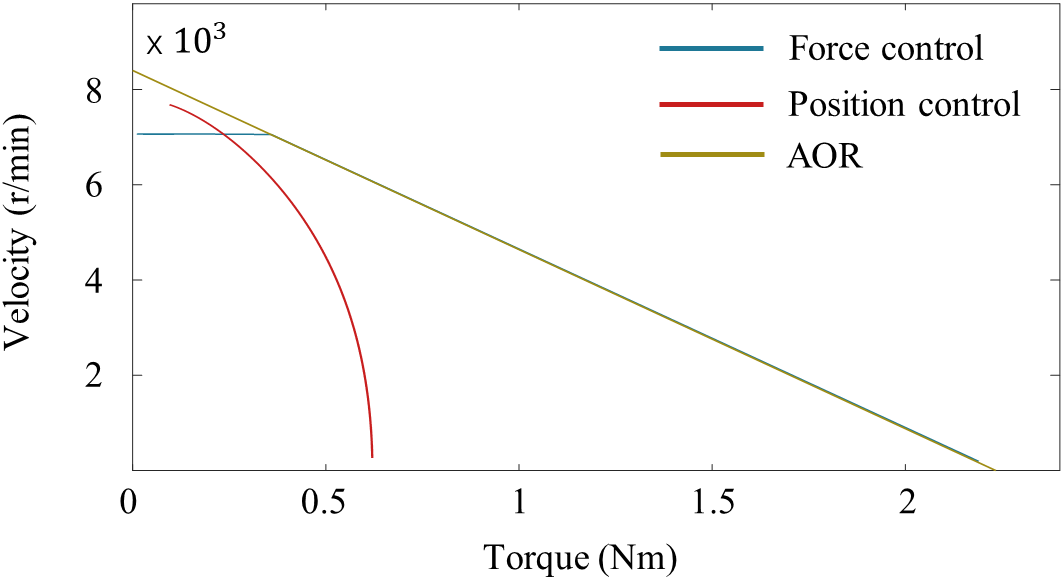}}
}
\caption{Simulation results for hopping leg. (a) the height of the foot end, (b) the actual saturation ratio of knee motor in the stance phase, (c) the $\dot\theta_{act}-\tau_{act}$ curve of the knee motor in stance phase.}
\label{fig:sim_graph}
\end{figure}

\subsection{Position Control}
As shown in Fig. \ref{fig:schematic}, the COG of body for two-mass model is connected to the COG of the single-legged robot and the mass point for foot is connected to the foot end of the robot. According to equation (\ref{eq:kin_ydes}), the control law of the actuators can be deduced by attitude calculation:
\begin{equation}
	\theta_{des}\left(t\right)=f\left(y_{des},t\right).
	\label{eq:con_angdes}
\end{equation}

\subsection{Simulation Verification}
To evaluate motor performance utilization in stance phase, the actual saturation ratio is defined as follows:
\begin{equation}
    C_{act}\left(t\right)=\frac{\left|\tau_{act}\left(t\right)\right|}{\left|\tau_{sat}\left(\dot\theta_{act},t\right)\right|}.
	\label{eq:sim_cact}
\end{equation}

For the sake of comparison, the average of $C_{act}$ can be figured out:
\begin{equation}
    C_{act\cdot avg}=\frac{1}{t_{lo}-t_{init}}\int_{t_{init}}^{t_{lo}} C_{act}\left(t\right) \text d t,
	\label{eq:sim_cactavg}
\end{equation}
where $t_{init}$ and $t_{lo}$ are the hopping time to start and to lift off.

The motion of a single-legged model is simulated with the ideal ground condition, where the foot during stance phase is fully grounded and does not generate any motion, namely, the ground can provide ideal ground reaction force. The system parameters utilized in the simulation are as shown in Table. \ref{tab:tabP} and Table. \ref{tab:tabF}.

\begin{table}[htbp]
	
	\setlength{\abovecaptionskip}{0cm}
	\setlength{\belowcaptionskip}{0cm}
	\caption{Force Control Parameters of the Hopping Leg}
	\centering
	\def\temptablewidth{0.45\textwidth}
	{\rule{\temptablewidth}{0.1pt}}  
	\begin{tabular*}{\temptablewidth}{@{\extracolsep{\fill}}crlrlc}
		\hline
		\hline
		&Parameter &Value&Parameter&Value&\\
		\hline
		&$m$&5.6 kg&$y_{s\cdot neu}$&0.45 m&\\
		\hline
		&$m_t$&1.87 kg&$C_{amp}$&0.12 m&\\
		\hline
		&$m_e$&0.8 kg&$C_{max}$&0.11&\\
		\hline
		&$k_s$&17 N/m&$k_p$&5424&\\
		\hline
		&$k_d$&9&&&\\
		\hline
		\hline
	\end{tabular*}
	{\rule{\temptablewidth}{0.1pt}}
	\label{tab:tabF}
\end{table}
\begin{table}[htbp]
	
	\setlength{\abovecaptionskip}{0cm}
	\setlength{\belowcaptionskip}{0cm}
	\caption{Position Control Parameters of the Hopping Leg}
	\centering
	\def\temptablewidth{0.45\textwidth}
	{\rule{\temptablewidth}{0.1pt}}  
	\begin{tabular*}{\temptablewidth}{@{\extracolsep{\fill}}crlrlc}
		\hline
		\hline
		&Parameter & Value &Parameter & Values \\
		\hline
		&$m$&5.6 kg&$y_{s\cdot neu}$&0.45 m&\\
		\hline
		&$m_t$&1.87 kg&$C_{amp}$&0.12 m&\\
		\hline
		&$m_e$&0.8 kg&$C_{max}$&0.11&\\
		\hline
		&$k_s$&17 N/m&&&\\
		\hline
		\hline
	\end{tabular*}
	{\rule{\temptablewidth}{0.1pt}}
	\label{tab:tabP}
\end{table}

As shown in Fig. \ref{fig:sim_graph} (a), the maximum value of the COG of body is 0.44 $m$ and 0.32 $m$ respectively for the position control and force control. In the simulation of force control, the saturation ratio of knee motor is close to 1 in the whole process (Fig. \ref{fig:sim_graph} (b)) and the value of $C_{act\cdot avg}$ is verified to be approximately 1 in stance phase, that means the performance of the knee motor has been maximized, while the saturation ratio of position control is much less than 1 and the average of saturation ratio $C_{act\cdot avg}$ is 0.5. In order to better understand the state of performance, we measured the $\dot\theta_{act}-\tau_{act}$ curve of the knee motor, as shown in Fig. \ref{fig:sim_graph} (c). The Admissible Operating Region (AOR) is formed by the maximal speed and torque points from the motor saturation model. The $\dot\theta_{act}-\tau_{act}$ curve of the force control is more closed to the AOR than that of the position control. The approach, force control with motor saturation model, is effective to increase saturation ratio and maximize foot clearance and performance of single leg.

\begin{figure*}[ht]
\centering
\setlength{\abovecaptionskip}{0.2 cm}
\setlength{\belowcaptionskip}{5 cm}
\includegraphics[width  = 6.8 in]{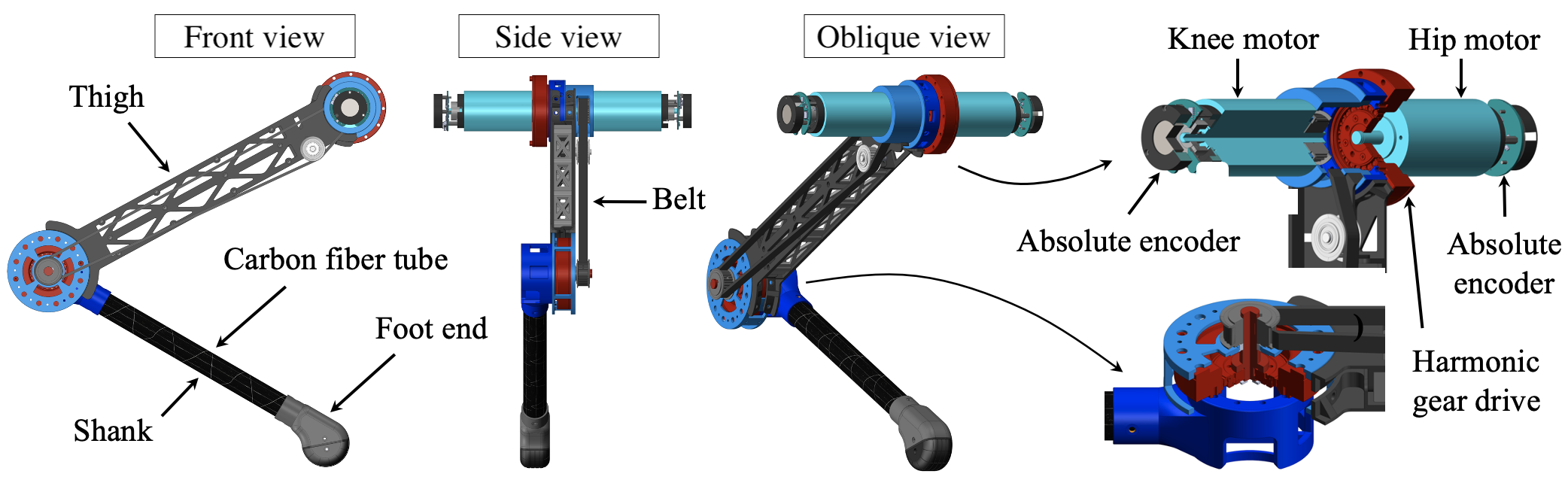}
\caption{Mechanical structure of the single leg. The left three subfigures are the front, side and oblique view of the single leg respectively. The right subfigure is the cross-sectional view of the hip (top) and knee joint (bottom) respectively.}
\label{fig:structure}
\end{figure*}
\section{Experiments}
Two experiments are illustrated in this section to verify that position control may reduce the force bandwidth of the control system and force control with motor saturation model can help actuator maximize the foot clearance and performance effectively in hopping experiment.

\begin{figure}[tbp]	
\centering
\centering
\subfigtopskip=0pt
\subfigbottomskip=6pt
\subfigcapskip=0pt
\vspace{-10pt}
\subfigure[]{
	\centering\includegraphics[width=3.0 in]{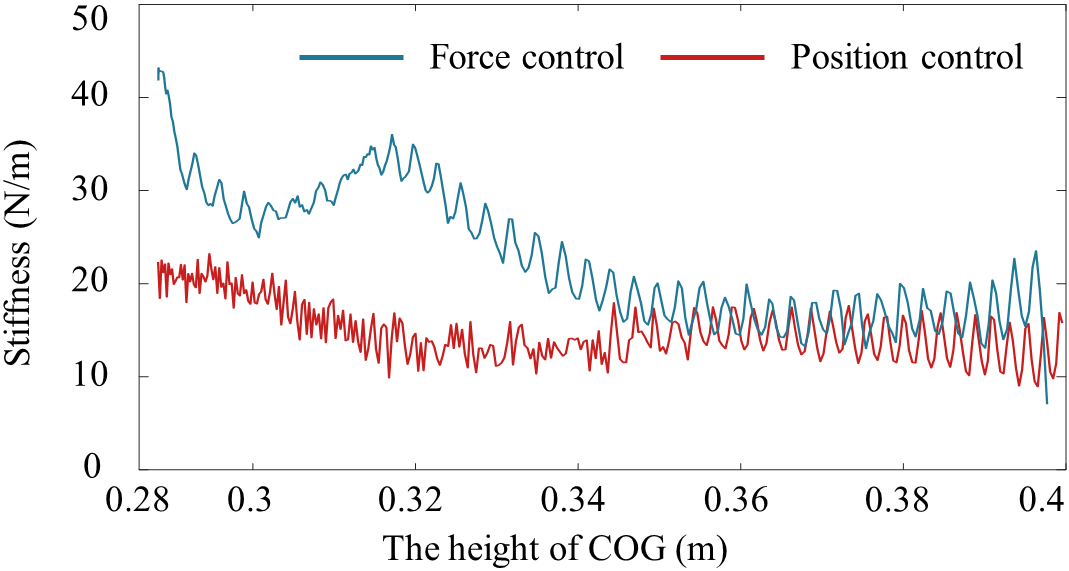} 	
}
\subfigure[]{
	\centering\includegraphics[width=3.0 in]{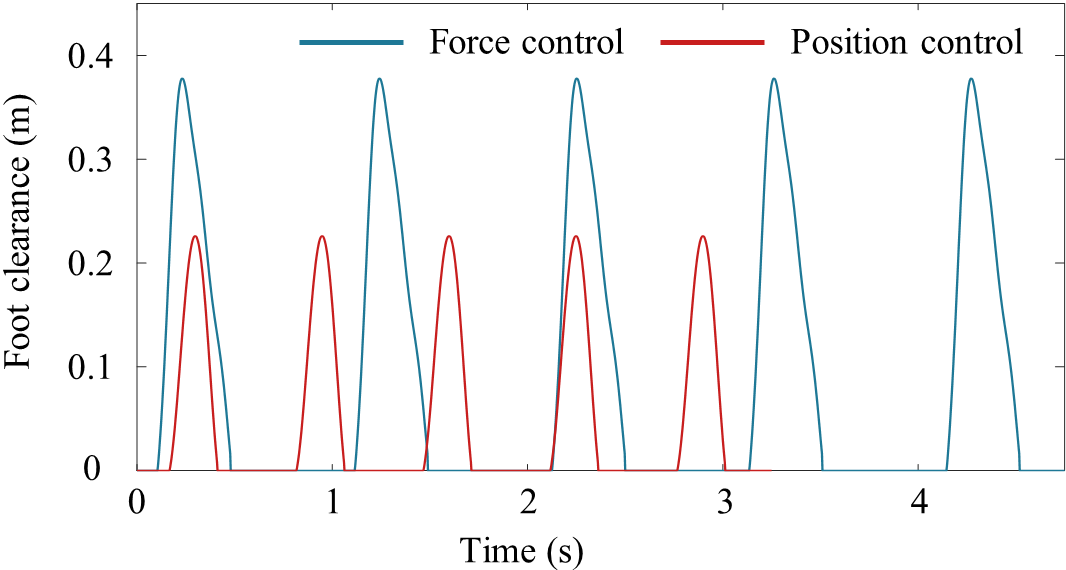}
}
\caption{Experimental results. (a) the actual stiffness $k_s$ of two-mass model for force control and position control in the stance phase, (b) the foot clearance for force control and position control.}
\label{fig:exp_sitff_clearance}
\end{figure}

\subsection{Experimental Platform}
As shown in Fig. \ref{fig:structure}, the single-legged robot platform has two joints, each joint consisting of a high-speed electric motor (Maxon EC45) equipped with a harmonic gear driven with reduction ratio of 100 and a 14 $bit$ absolute encoder. In order to minimize the inertia of the robot’s leg, both the motors are mounted on the hip joint. Different from the hip joint in dynamic structure, the knee joint is driven by a motor mounted on the hip through a synchronous belt from which the torque is transmitted to the harmonic gear. The robot platform has a thigh length of 0.38 $m$ and a shank length of 0.361 $m$, which is equipped with horizontal and vertical guides, so that the robot can move forth or back, up or down. The high reduction ratio of harmonic gear and synchronous belt make the structure absorb the torque fluctuation of the motor input more efficiently. The control system use a Windows computer as EtherCAT's host. EtherCAT host on Windows is based on TwinCAT and runs at 4 $khz$.

\begin{figure}[tbp]	
\centering
\centering
\subfigtopskip=0pt
\subfigbottomskip=6pt
\subfigcapskip=0pt
\vspace{-10pt}
\subfigure[]{
	\centering{\includegraphics[width  = 3.2 in]{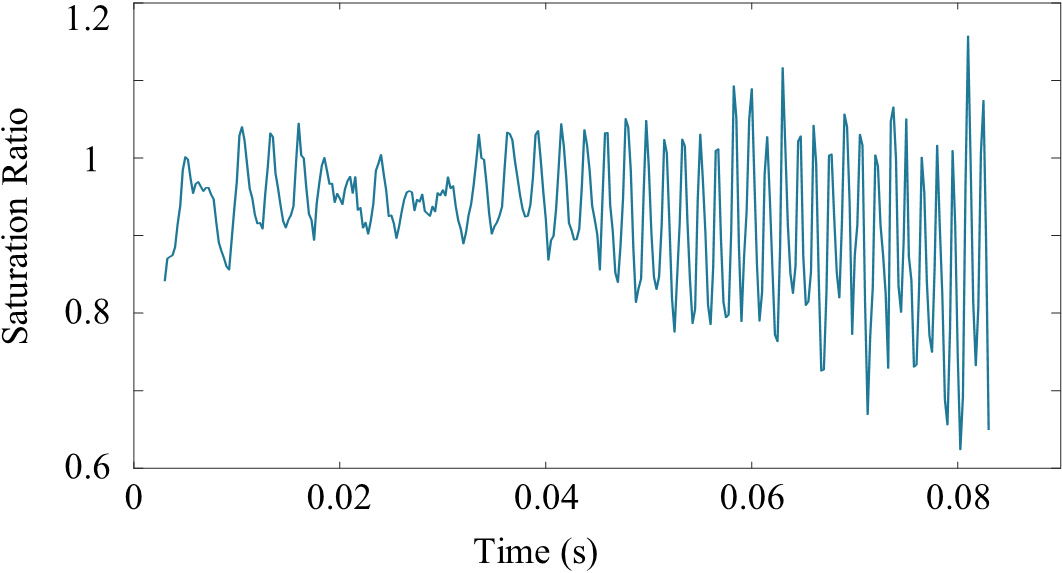}} 	
}
\subfigure[]{
	\centering{\includegraphics[width  = 3.2 in]{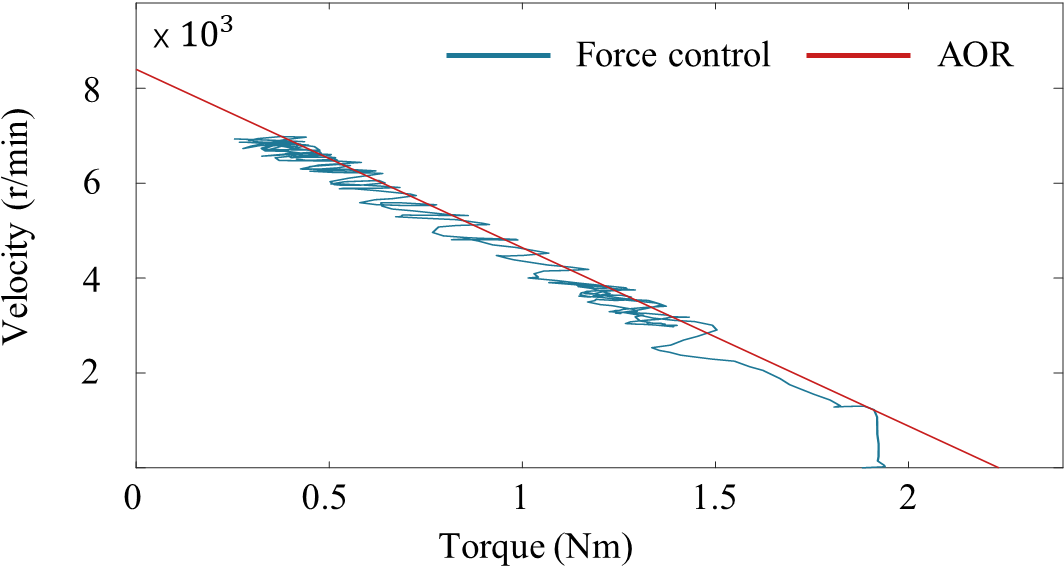}}
}
\caption{Experimental results for force control experiment. (a) the actual saturation ratio of knee motor in stance phase, (b) the $\dot\theta_{act}-\tau_{act}$ curve of the knee motor in stance phase.}
\label{fig:exp_f_graph}
\end{figure}
\begin{figure}[tbp]	
\centering
\centering
\subfigtopskip=0pt
\subfigbottomskip=6pt
\subfigcapskip=0pt
\vspace{-10pt}
\subfigure[]{
	\centering{\includegraphics[width  = 3.2 in]{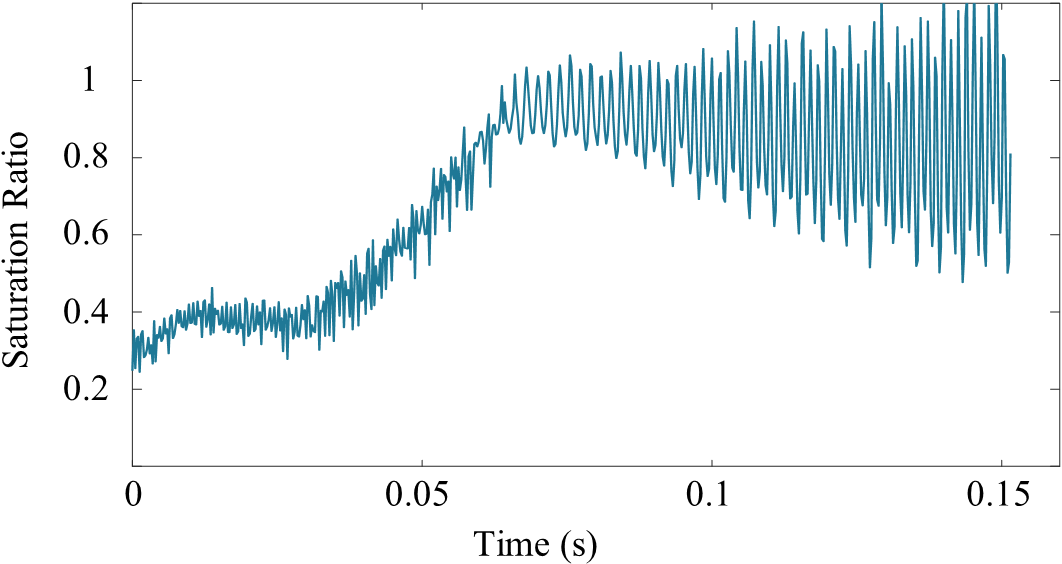}} 	
}
\subfigure[]{
	\centering{\includegraphics[width  = 3.2 in]{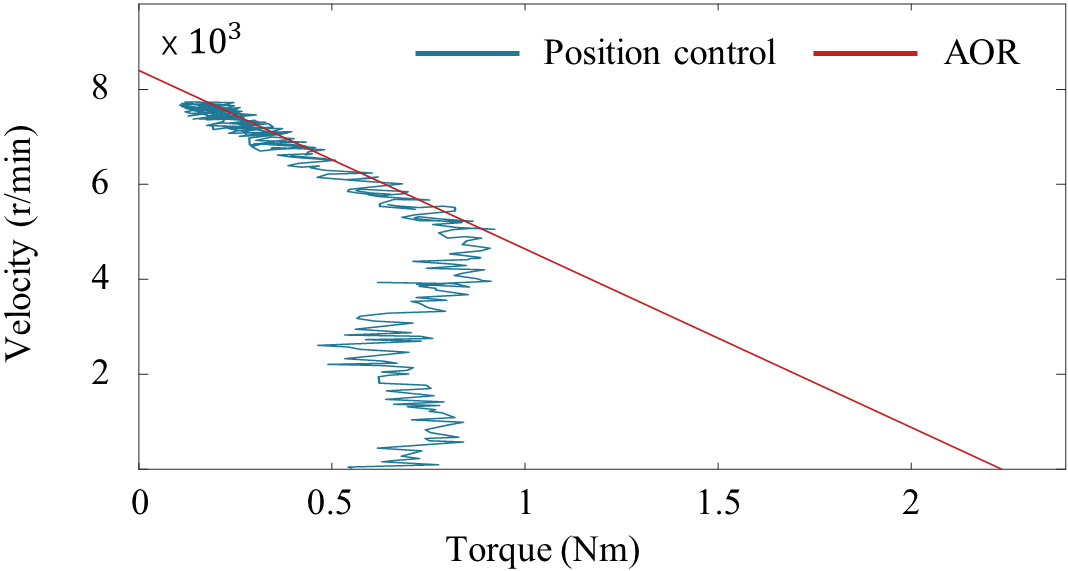}}
}
\caption{Experimental results for position control experiment. (a) the actual saturation ratio of knee motor in stance phase, (b) the $\dot\theta_{act}-\tau_{act}$ curve of the knee motor in stance phase.}
\label{fig:exp_p_graph}
\end{figure}

\subsection{Force Control Experiment}
We set the same initial height of $h$. The system parameters utilized in this experiment are as shown in Table. \ref{tab:tabF}. 

As shown in  Fig. \ref{fig:exp_sitff_clearance} (a), the actual stiffness of the two-mass model for the force control is larger than that for position control at the height of COG from 0.28 $m$ to 0.4 $m$. According to Fig. \ref{fig:exp_sitff_clearance} (b) and Fig. \ref{fig:exp_picture}, the maximum value of $h_r$, the height of foot end, is 0.38 $m$ in force control experiment, which is almost 58 percent higher than the result of the position control experiment. The extreme performance of the actuator is significantly improved. According to the experimental results in Fig. \ref{fig:exp_f_graph} (a), the saturation ratio of the knee motor is close to 1 during the whole stance phase, indicating that the motor performance has been maximized. To verify the results further, we measured the $\dot\theta_{act}-\tau_{act}$ curve of the knee motor in stance phase, as shown in Fig. \ref{fig:exp_f_graph} (b). No matter at high or low speeds, the $\dot\theta_{act}-\tau_{act}$ curve has coincided with the AOR curve and $C_{act \cdot avg}$ is also close to 1.

\subsection{Position Control Experiment}

\begin{figure*}[ht]
\centering
\setlength{\abovecaptionskip}{0.3 cm}
\setlength{\belowcaptionskip}{3 cm}
\includegraphics[width = 6.6 in]{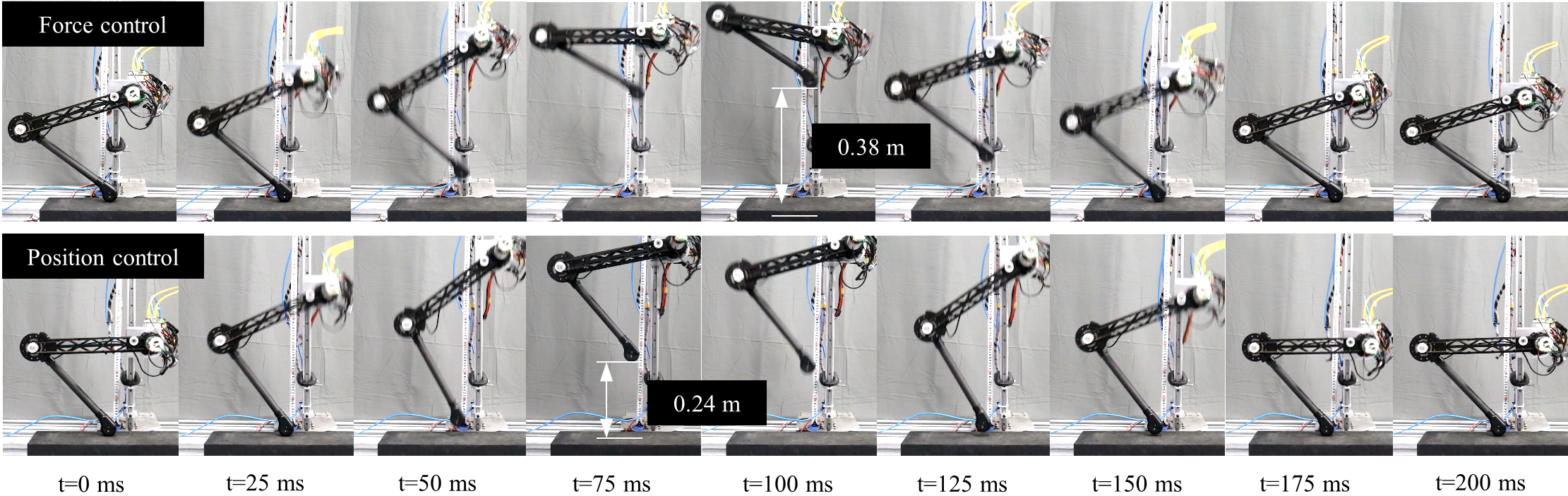}
\caption{The screenshot of the hopping experiment on the single leg with position control and force control respectively. The screenshot is taken at every 40 ms. From the comparison, hopping height of the foot with the force control is higher due to maximizing the motor power.}
\label{fig:exp_picture}
\end{figure*}

Through two-mass model, the change law between the COG and the foot end is deduced, so do the control law of the hip actuator and the knee actuator. With the position accuracy guaranteed, we set the same initial attitude and maximized the spring stiffness to test the extreme performance of the hopping. The system parameters utilized in this experiment are shown in Table. \ref{tab:tabP}.

As shown in  Fig. \ref{fig:exp_picture}, the maximum value of $h_r$ is 0.24 $m$, which is much smaller than the height obtained in the second experiment. As can be seen from Fig. \ref{fig:exp_p_graph} (a), although the spring stiffness is maximized, the saturation ratio of the knee motor is still far less than 1 in the first half period of the stance phase. The value of $C_{act \cdot avg}$ is 0.7, which is still 30 percent lower than 1. In fact, the larger force the actuator outputs during that time, the higher the robot can jump. To find out more about the problem, we measured the $\dot\theta_{act}-\tau_{act}$ curve of the knee motor, as shown in Fig. \ref{fig:exp_p_graph} (b). It can be seen from the figure that the motor tends to be saturated in the high speed section, but there is still a large margin of motor performance in the low speed section. In conclusion, in the primary stage of stance phase, the actuator output is insufficient, leading to poor hopping performance. Firstly, under the specified acceleration, there is an inertia error between the control model and the actual structure, which leads to the fact that sometimes the actuator needs to output the reverse torque to compensate for the error. Secondly, when position errors are generated and accumulated, the actuator must trade high utilization of performance for position accuracy. Thirdly, as mentioned at the beginning, low stiffness and high elasticity will reduce the force bandwidth of the control system.

The experiments for the hopping task demonstrates that the method proposed in this study is able to accomplish stable and continuous hopping task, which has the potential to be applied in the legged locomotion such as biped and quadruped locomotion.
\section{Conclusion}
This letter proposes a control method considering the motor saturation for a hopping leg that is able to accomplish the maximum foot clearance. Motivated by the phenomenon, in which legged animals tends to fully retract their legs so as to hop over an obstacle as high as possible, this study explores a method to make best use of the capacity of the electric motor to maximize the foot clearance. This method is based on the force control of the joints and a two-mass model. Accurate motor saturation model is analyzed to guarantee the effectiveness of force control. A robotic leg with two joints and actuated by electric motor is designed for experimental validation. The joint actuator includes electric motor and harmonic gear, which are able to improve the power density. Through the maximizing the desired torque and the saturation ratio, the foot clearance and the performance of the hopping leg are further leveraged. Benefiting from the proposed method, the performance of the actuator is no longer limited to the traditional control strategy for the force and position control, which tends to reduce the force bandwidth of the control system. This study demonstrates in experiments that the proposed method is able to support the continuous and stable hopping, which is proven be effective in a long-term working condition for the hopping task in the legged locomotion. Future work include implementation of the hopping control method on a quadruped robot for the dynamic locomotion.

\bibliographystyle{elsarticle-num-names} 
\bibliography{references}

\end{document}